\documentclass[11pt]{article}

\usepackage[preprint]{acl}
\usepackage{adjustbox}
\usepackage{times}
\usepackage{latexsym}
\usepackage{amsfonts}
\usepackage{amssymb}
\usepackage{multirow}
\usepackage{graphics}
\usepackage{tabularx}
\usepackage{arydshln}
\usepackage{array} 
\usepackage{url}

\usepackage[T1]{fontenc}

\usepackage[utf8]{inputenc}

\usepackage{microtype}

\usepackage{inconsolata}

\title{Team QUST at SemEval-2024 Task 8: A Comprehensive Study of
Monolingual and Multilingual Approaches for Detecting AI-generated Text}

\author{Xiaoman Xu, Xiangrun Li, Taihang Wang, Jianxiang Tian, Ye Jiang \\
        College of Information Science and Technology   \\  
        Qingdao University of Science and Technology, China  \\
        }
        
\begin{document}
\maketitle
\begin{abstract}
This paper presents the participation of team QUST in Task 8 SemEval 2024. We first performed data augmentation and cleaning on the dataset to enhance model training efficiency and accuracy. In the monolingual task, we evaluated traditional deep-learning methods, multiscale positive-unlabeled framework (MPU), fine-tuning, adapters and ensemble methods. Then, we selected the top-performing models based on their accuracy from the monolingual models and evaluated them in subtasks A and B. The final model construction employed a stacking ensemble that combined fine-tuning with MPU. Our system achieved 8th (scored 8th in terms of accuracy, officially ranked 13th) place in the official test set in multilingual settings of subtask A. We release our system code at:\url{https://github.com/warmth27/SemEval2024_QUST} 
\end{abstract}

\section{Introduction}

Large language models (LLMs) enable quick, coherent responses and content creation but also raise ethical concerns about misinformation and academic integrity \cite{wang2023m4}. To differentiate between machine-generated and human-created content, previous study \cite{guo2023authentigpt} has been extensively discussed in industry and academic works. 

Semeval 2024's Task 8 \cite{semeval2024task8} encourages the participants to develop an automatic system for detecting AI-generated text by leveraging an extended version of the M4 dataset \cite{wang2023m4,wang2024mg-bench}. We engaged in subtasks A and B, during which we encountered the challenges of overcoming linguistic differences, data scarcity, and inadequate cross-lingual generalization capabilities. Furthermore, existing multilingual models are less discussed in detecting AI-generated text, compared to monolingual ones, which further exacerbates the difficulty in model selection. 

\begin{figure}[h!]
	\centering
		\includegraphics[scale=0.35]{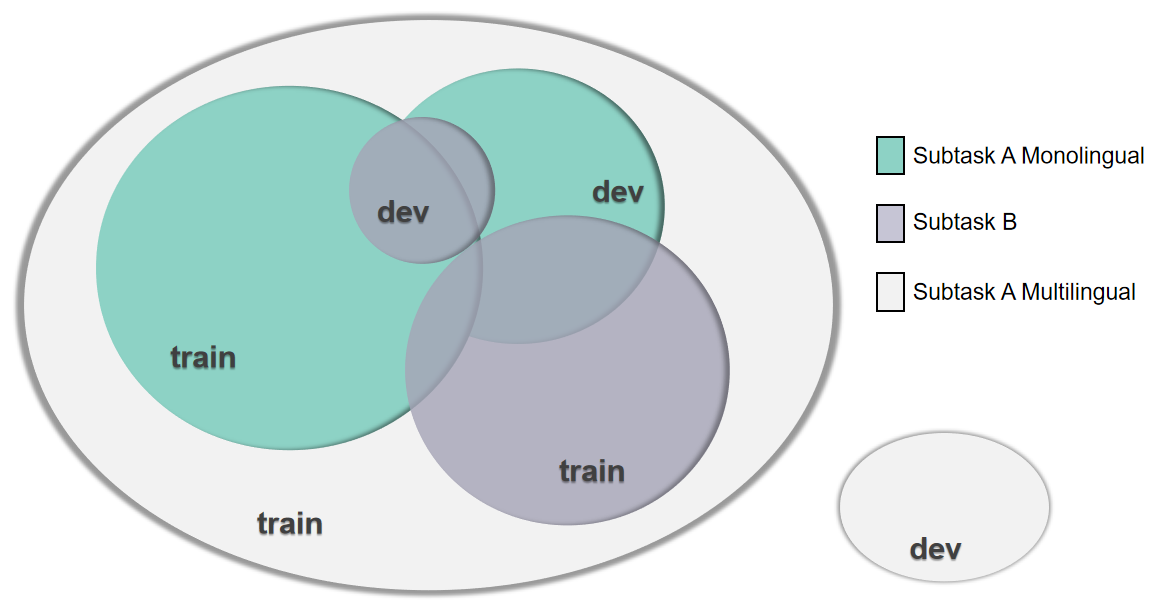}
	\caption{The data distribution in subtask A and B.}
	\label{FIG:fake}
\end{figure}

Meanwhile, we found that the multilingual dataset in subtask A contains data from both monolingual data and subtask B, as shown in Figure \ref{FIG:fake}. To enhance the diversity and scale of the text dataset, we performed back-translation on the multilingual training set to increase the volume of monolingual data and conducted data cleaning to improve data quality.

In our approach to subtasks A and B, we initially applied deep learning methods for a swift assessment in subtask A and proceeded to fine-tune multiple pre-trained language models (PLMs), inspired by recent studies highlighting the efficacy of fine-tuning methods in text classification. However, the training cost of the fine-tuning method is relatively high. We further utilize the Adapter \cite{hu2021LoRA} to parameter efficiency fine-tune (PEFT) \cite{hu2021LoRA} the model while preserving its performance. We also noted that despite the reduced training time, the performance of adapter models was not consistently stable, showing variability across experiments. Given the critical importance of model performance, decided not to utilize adapters in the testing phase.

To enhance model performance and generalizability, we adopted a stacking-based ensemble learning method, utilizing the logits from the top two performing models as inputs for a linear layer to generate final predictions. Finally, our experimental results on the test set show that integrating data augmentation and ensemble learning significantly improves model efficacy in task-specific settings.

\section{System description}

\subsection{MPU framework}

Recent research on machine-generated text recognition has evolved into treating it as a binary classification problem, with the latest advancements including the Multiscale Positive-Unlabeled (MPU) \cite{tian2023multiscale} training framework. This approach introduces a length-sensitive MPU loss combined with abstract recurrent models and a text multi-scale module, significantly enhancing detection performance for short texts.

Upon analyzing the text lengths in official datasets, As shown in table \ref{count}, we observed a predominance of short texts, with those exceeding 512 characters making up a quarter of the total. This insight highlighted the MPU framework's suitability for subtask A. The MPU model, previously tested only on the HC3 \cite{guo-etal-2023-hc3} Chinese and English datasets, needed assessment for its effectiveness on multilingual datasets. To address this, we integrated the MPU framework with the XLM-Roberta (XLM-R) model to enhance its adaptability for multilingual tasks and employed stacking ensemble techniques, yielding significant improvements in our experimental outcomes.

\subsection{Fine-tuning}
Fine-tuning PLMs such as BERT or RoBERTa have been extensively discussed in text classification tasks \cite{jiang2023team,jiang2023similarity,jiang2020comparing}. Recently, the DeBERTa model \cite{he2020deberta} is an enhancement built upon the foundations of BERT and RoBERTa through the incorporation of a disentangled attention mechanism and an enhanced masked decoder. We utilized the DeBERTa model and performed fine-tuning on it in our experiment. Although fine-tuning PLMs to specific domains or downstream tasks is a crucial and common practice, fully fine-tuning its large number of parameters becomes time-consuming and costly.

\subsection{Adapter}
In our experiments, due to the substantial size of the DeBERTa model and the size of the official datasets, each fine-tuning run required a significant amount of time. Adapter-based fine-tuning is an approach to fine-tuning a PLM that involves freezing the most of layers and inserting low-dimensional adapter modules into each layer to improve parameter efficiency. Research has shown that introducing adapters reduces the number of trainable parameters to 3.6\%, with only a marginal performance drop of 0.4\% \cite{houlsby2019parameter}. Furthermore, in some cases, models applying adapters perform even better \cite{bapna2019simple}.

In our task, we employed the LoRA (Low-Rank Adapter) method \cite{hu2021LoRA}, which injects trainable rank-decomposition matrices into each layer of the Transformer architecture, effectively freezing the PLM's weights. This significantly reduces the number of trainable parameters for downstream tasks. Therefore, we added a sequence classification head on top of the model to adapt the PLMs to the classification task. This has reduced the training costs and shortened the training time.

\subsection{Stacking}

Ensemble learning combines multiple base learners to form a predictive model with enhanced generalization capabilities \cite{sagi2018ensemble}.  Initially, predictions are generated employing various machine learning algorithms. Then, these predictions serve as inputs for a subsequent classifier. Upon training the subsequent classifier, the integrated model is optimized to produce a new prediction set.

\section{Experimental setup}
\subsection{Data preprocessing}
The subtasks A and B involving diverse domains and sources with both human and machine-generated texts, we encountered chaotic symbols and extraneous content such as hyperlinks, numerals, and escape characters. To improve data quality, we undertook preprocessing steps including: removing special characters; eliminating excessive whitespace and line breaks; discarding Unicode escape characters and numerically formatted texts; removing hyperlinks; excluding irrelevant text lines like those for sharing, surveys, comments, ads, terms of use, and copyright notices; and deleting duplicate sentences. Notably, we avoided removing escape characters from multilingual training and validation sets to preserve original characters in non-English texts.

\begin{table*}
\centering
\begin{tabular}{l|cc|cc|ccc}
\hline
\textbf{\multirow{2}{*}{Subtask}} & \multicolumn{2}{c}{\textbf{v1}}  & \multicolumn{2}{c}{\textbf{v2}} & \multicolumn{3}{c}{\textbf{v3}}   \\ \cline{2-8}

                         & train       & dev      & train       & dev      & train   & dev   & test   \\
                         \hline
\textbf{Monolingual}              & 119,757     & 5,000    & 167,252     & 5,000    & 176,252 & 5,000 & 34,272 \\
\textbf{Multilingual}             & 172,417     & 4,000    & 172,417     & 4,000    & 176,417 & 4,000 & 42,378 \\
\textbf{subtask B}                & 71,027      & 3,000    & 105,908     & 3,000    & 176,252 & 3,000 & 18,000\\
\hline
\end{tabular}
\caption{The overall data statistic. "v1" and "v2" respectively refer to the original dataset, and the dataset processed after data augmentation. "v3" refers to the training dataset that has undergone data augmentation and other processing after the official test dataset was released.}
\label{data count}
\end{table*}

\begin{table*}
\centering
\begin{adjustbox}{max width=1\textwidth}
\begin{tabular}{l|ccc|ccc|ccc}
\hline
 \textbf{Statistic} & \multicolumn{3}{c}{\textbf{SubA-mono}} & \multicolumn{3}{c}{\textbf{SubA-multi} }& \multicolumn{3}{c}{\textbf{SubB}}
 
 \\ \cline{2-10}

 & Train & Dev & Test & Train& Dev & Test& Train & Dev & Test \\
 \hline

\textbf{avg-sent }           & 16.5        & 13.0 & 18.4 & 15.7  & 8.9  & 17.1  & 15.3    & 10.2    & 17.5   \\

\textbf{avg-sent-len}         & 24.7        & 26.8 & 23.7 & 25.5   & 22.9 & 23.1   & 23.5    & 24.1   & 23.6  \\

\textbf{sent-len\textgreater{}512} & 16.7\%       & 13.1\% & 23.5\%  & 17.3\%   & 1.2\%   & 14.6\%   & 15.2\%   & 6.6\%    & 18.9\%   \\
\hline
\end{tabular}
\end{adjustbox}
\caption{The statistics of the v2 dataset.
"avg-sent" represents the average number of sentences per document, "avg-sent-len" represents the average number of words per document, "sent-len\textgreater{}512" represents the percentage of documents that their sent-len are greater than 512 words.}
\label{count}
\end{table*}

\subsection{Data augmentation}

We evaluated our models on the original dataset (\textbf{v1}) before the test set was released. We found that the multilingual set contained training and validation data for the monolingual and subtask B. To enlarge the monolingual dataset and improve model performance, we removed 5000 monolingual validation entries from the multilingual set, translated Chinese, Indonesian, Urdu, and Bulgarian data to English using Google API, and then cleaned the data to produce a refined dataset (\textbf{v2}).

we calculated the statistics of different versions of datasets, as shown in Table \ref{data count}. For the multi-class subtask B, we re-labeled the dataset based on multilingual tags, addressing a severe imbalance by reducing instances in overrepresented categories for balance. After receiving the test set, we included the multilingual validation set into our training dataset and performed the same enhancement processes, creating a \textbf{v3} version for final model training and prediction.

We further analyzed the augmented version of the v2 dataset, as shown in Table \ref{count}. This analysis includes the average sentence length and average text length, as well as the proportion of texts exceeding 512 characters in length. The average sentence count was obtained through sentence tokenization using the sent-tokenize tool, while the average sentence length was calculated using the word-tokenize tool from the NLTK python library. Model performance for subtasks A and B was evaluated based on accuracy.

\subsection{Monolingual models}

In subtask A, submissions were made using two systems based on the different language tracks. For the monolingual English track, the system consisted of five approaches across ten models, as detailed in Table \ref{result}. The final submission system employed a stacking ensemble method, which was composed of the two best-performing models out of the ten.

In Table \ref{result}, these models had a learning rate of 1e-4, were trained for 3 epochs, and the best-performing models on the validation set were saved. For the fine-tuned models, we adhered to the inherent 512-token length limitation to ensure consistency in the input data and effective processing by the models.

The final monolingual model selected the top-performing two models, DeBERTa-v3-large and RoBERTa-base model based on MPU framework(RoBERTa-base-MPU), for stacking ensemble learning. The learning rate for the ensemble model remained set at 1e-4, trained for 1000 epochs. Only the best-performing stacking model was retained, and the final predictions were based on this optimal stacking model.

\subsection{Multilingual models} 

In the multilingual track, we conducted experiments employing the top five models that exhibited promising performance in monolingual contexts. We opted to substitute the RoBERTa model with the XLM-R model, which is specifically designed for multilingual tasks. 

Derived from the RoBERTa architecture, the XLM-R model has undergone training across 100 distinct languages, endowing it with multilingual capabilities. This versatility enables the model to process and comprehend various languages effectively, leading to notable enhancements in performance across diverse cross-lingual transfer tasks. Subsequently, we identified the top two models for integration through stacking ensemble. The integration of predictions from these two models yielded superior predictive performance.

Following this, we trained the multilingual models on the v2 version of the multilingual training dataset. We set the learning rate to 1e-4 and 3 epochs while retaining the models that performed best on the validation set.

During the experimentation, we observed that the addition of adapters to the DeBERTa-v3-large model resulted in unstable performance, while direct fine-tuning of the DeBERTa-v3-large model exhibited better results. Based on this observation, we ultimately chose to integrate the XLM-R-MPU model and the DeBERTa-v3-large model. To achieve this, we saved the best models from each training session and utilized these two optimal models to generate logits. Subsequently, we merged the logits from both sets of models as part of the training data for the stacking ensemble's input linear layer. We set the learning rate to 1e-4 and extended the training epochs to 1000 to ensure thorough model training. Throughout this process, we continuously monitored and retained the best-performing stacking model, which was ultimately applied to the test set for final predictions.

\subsection{Subtask B models}

We extended the binary classification capabilities of the RoBERTa-base model combined with the MPU method to address multi-class problems in subtask A, employing a one-vs-rest strategy for six categories including human, ChatGPT, etc. This resulted in six separate classifiers, with classification based on the highest confidence level among positive predictions from these classifiers for each category.

Following this, we opted for the consistently excellent performance of the LoRA adapter-based DeBERTa-v3-large (DeBERTa-v3-Large-LoRA) model and applied it to subtask B. Additionally, we introduced a new adapter-based RoBERTa-large model. The model configurations were consistent with the monolingual models. In the final ensemble model, we employed stacking with the DeBERTa-v3-Large-LoRA and RoBERTa-large models. The learning rate was set to 1e-4, with a training period of 3000 epochs, and only the model with the highest score was retained.

\section{Results}

\begin{table*}[ht]
\centering
\begin{tabular}{llcccc}
\hline
\textbf{Methods} & \textbf{Models} & \textbf{Mon1} & \textbf{Mon2} & \textbf{Mul} & \textbf{SubB}\\
\hline
\textbf{\multirow{4}{*}{Deep learning}} & CNN \cite{jiang2019team}  & 0.762 & - & - & - \\
                                & RNN \cite{lin2017structured} & 0.729 & - & - & - \\
                                & RCNN \cite{lin2017structured} & 0.702 & - & - & - \\
                                & Self-Attention \cite{jiang2023topic} &0.762 & - & - & - \\
\hline
\textbf{\multirow{2}{*}{MPU}}           & RoBERTa-base-MPU \cite{tian2023multiscale}  & 0.894 &  0.979 & - & - \\
                               &XLM-R-MPU (Ours) & - & - & \textbf{0.798} & 0.7\\
\hline
\textbf{\multirow{6}{*}{Fine-tuning}}   & DeBERTa-v3-base \cite{he2021debertav3} & 0.823 & - & - & - \\
                                & DeBERTa-v3-large \cite{he2021debertav3} & 0.84 & 0.979 & 0.763 & -  \\
                                & longformer-base-4096 \cite{Beltagy2020Longformer} & 0.737 & - & -  & - \\
                                & BERT \cite{devlin2018bert} & 0.769 & 0.955 & 0.654 & - \\
                                & RoBERTa-base \cite{DBLP:journals/corr/abs-1907-11692} & -  & -    & -  & 0.75\\
                                & XLM-R \cite{DBLP:journals/corr/abs-1907-11692} & -  & -    & 0.72  & -\\
                                
\hline
\textbf{\multirow{2}{*}{Adapter}}       & DeBERTa-v3-Large-LoRA (Ours) & 0.843 & 0.948 & 0.669 & 0.858\\
                               & Roberta-large-LoRA (Ours) & -   & -  & -  & 0.862\\
\hline
\textbf{\multirow{2}{*}{Stacking}}     & RoBERTa-base-MPU+DeBERTa-v3-large (Ours) & \textbf{0.96} & \textbf{0.99} & 0.795 & -\\
                              & RoBERTa-large+DeBERTa-v3-large (Ours) & - & - & - & \textbf{0.94}\\

\hline
\end{tabular}
\caption{The overall accuracy comparison in subtask A and B. "Mon1", "Mul", and "SubB" respectively represent the accuracy of monolingual models, multilingual models, and subtask B models trained on the v1 dev set. "Mon2" is the dev accuracy on the v2 dataset.}
\label{result}
\end{table*}

\subsection{Monolingual results}

Our evaluation work is divided into two main stages: first, the evaluation based on the officially provided monolingual dataset (\textbf{Mon1}), and second, the evaluation based on our back-translated and processed monolingual dataset (\textbf{Mon2}).

In the \textbf{\textcolor{black}{Mon1}} evaluation stage, we aimed for a quick baseline model implementation, using traditional deep-learning methods. The results in Table \ref{result} on the Mon1 dataset show that traditional models generally outperformed the fine-tuned deep learning models, likely due to the small size of the official monolingual dataset, which contains only about 110,000 entries. Consequently, large models like RoBERTa-base may not be adequately trained, while smaller models such as CNN or RNN could perform better by being less prone to overfitting.

By integrating the MPU framework with the RoBERTa-base model, performance improved by 20 percentage points over direct fine-tuning, highlighting MPU's benefits in boosting short-text performance and enhancing machine-generated long text detection. Despite being designed for long documents, Longformer-base-4096 underperformed compared to CNN and Self-Attention methods on the Mon1 dataset.

The DeBERTa model, an advancement over BERT and RoBERTa, excelled in our tests, especially after fine-tuning with adapters, which improved both efficiency and performance, slightly surpassing the fully fine-tuned DeBERTa. Stacking and re-predicting logits from the top two models led to a nearly 7\% improvement over the best single model, underscoring the effectiveness of model fusion in increasing prediction accuracy and stability.

In the \textbf{\textcolor{black}{Mon2}} evaluation stage, we retrained them on the Mon2 dataset after selecting the top five best-performing models on the Mon1 dataset. After retraining, the performance of the models on the Mon2 dataset improved by approximately 20\%. likely due to shorter texts enhancing feature detection, noise reduction from removing poor-quality data, and increased dataset diversity and size. These factors combined allowed the models to gain a deeper understanding of language characteristics.

\subsection{Multilingual results}

Experiments performed on a refined multilingual dataset employing the DeBERTa-v3-Large-LoRA model produced a performance score of merely 0.669, notably inferior to the baseline model's performance on the unprocessed dataset. This discrepancy may stem from improperly removing crucial features during the dataset cleaning process or introducing errors. Therefore, we opted to directly train the selected model on the raw official multilingual dataset, as detailed in Table \ref{result} above. We found that the XLM-R model integrated with the MPU framework outperformed the baseline XLM-R model by 7\% on the dataset, thus confirming the effectiveness of the MPU framework.

While the BERT model excels in monolingual tasks, its performance lags behind the baseline by approximately 7\% in multilingual tasks, suggesting that BERT may be less suitable for multilingual classification tasks. The DeBERTa-v3-large model, which is an improvement based on BERT and RoBERTa, outperforms the baseline XLM-R by 3.55\% on multilingual datasets. This improvement can be attributed to DeBERTa's optimizations to both architectures, which prove particularly effective in multilingual processing, enhancing the model's learning capabilities and generalization.

In our experimental results table, we observe that the performance of the DeBERTa-v3-Large-LoRA  model is 5.1\% lower than the baseline model, while also exhibiting a 9.4\% decrease compared to directly fine-tuning the DeBERTa-v3-large model. This discrepancy in performance may stem from significant differences in data distribution between the pre-training task and incremental training. 

Specifically, there exists a substantial difference in data distribution between the DeBERTa model and the model fine-tuned via LoRA adapters, resulting in insufficient parameter updates to effectively capture these differences. This phenomenon suggests that although LoRA adapters offer a parameter-efficient fine-tuning method, relying solely on limited parameter adjustments may not suffice to achieve optimal performance in situations with substantial disparities in data distribution.

The stacking results in the multilingual task failed to surpass the performance of the XLM-R-MPU model, which could be attributed to the already robust nature of XLM-R-MPU, potentially causing the ensemble model to overfit on the training data, thereby reducing performance on validation or test data. Another possibility is that the two top-performing models exhibit high correlation in predictions (i.e., commonly making the same type of errors), thus stacking them may not yield significant performance improvements.

\subsection{Subtask B results}

Table \ref{result} indicates that the performance of the XLM-R-MPU model continues to deteriorate on the dataset for subtask B, indicating poor results. This could be attributed to the model originally being designed for binary classification tasks and not being well-suited for multi-class tasks. 

We found that by directly freezing the model and fine-tuning the adapter-based DeBERTa-v3-Large (DeBERTa-v3-Large-LoRA) and RoBERTa-large (Roberta-large-LoRA) models, the classification effectiveness significantly improved, outperforming the official baseline by approximately 10 percentage points.The use of the LoRA adapter allows models to more effectively utilize pre-trained knowledge while avoiding over-fine-tuning and reducing the risk of overfitting on specific tasks. After data cleaning, the pre-trained data used by DeBERTa-v3-large and RoBERTa-large were closer to the target multi-class task, potentially further enhancing their performance.

\section{Ablations}
We conducted ablation experiments using the DeBERTa-v3-large and Roberta-base models on the Mon1 dataset. As shown in Table \ref{Ablations}, the experimental results indicate that the introduction of the MPU framework and stacking ensemble method significantly improves the model's performance, resulting in a notable performance enhancement

However, despite the inclusion of the LoRA adapter, the performance improvement is not significant. This could be attributed to the insufficient number of parameters fine-tuned solely by the adapter when faced with complex tasks, which hinders the model from learning additional knowledge effectively.

\begin{table}[h]
\centering
\begin{tabular}{ll}
\hline
\textbf{Methods}         & \textbf{Results} \\ \hline
\textbf{DeBERTa-v3-large}         & 0.84             \\
\textbf{DeBERTa-v3-large  w/  LoRA} & 0.843            \\
\textbf{Roberta-base}         & 0.694            \\
\textbf{Roberta-base  w/  MPU}  & 0.894            \\
\textbf{Stacking}        & \textbf{0.96}    \\ \hline
\end{tabular} 
\caption{Ablation experiments. "Stacking" refers to the aggregation of results from the "DeBERTa-v3-large with LoRA" model and the "Roberta-base with MPU" model.}
\label{Ablations}
\end{table}

\section{Official test results}
Our system ranks 8th on Semeval 2024 Task 8 official multilingual test set of subtask A.  It is noteworthy that, employing the same method, Only the model trained on the multilingual dataset surpassed the baseline of 0.80 with a score of 0.90, while models on monolingual datasets and Task B exhibited comparatively inferior performance, with none of the submissions reaching the respective domain baselines. 

This phenomenon could be attributed to several factors. Firstly, although monolingual validation sets were removed from the multilingual training set, the processed multilingual training data may still share similarities with monolingual validation sets. This could lead to superior model performance during evaluation; however, due to disparities between the final test set data and the processed multilingual data, models may fail to meet baseline performance on the test set. Secondly, model training based on fine-tuning may encounter instability and catastrophic forgetting issues, thus affecting the model's generalization ability. Therefore, even with the same model, discrepancies in performance on official test sets may arise due to differences between datasets, resulting in significant performance gaps.

\section{Conclusion}

In conclusion, our team developed three distinct systems for SemEval-2024 Task 8, targeting the monolingual and multilingual aspects of subtask A and addressing subtask B. We achieve 8th place in the multilingual setting of subtask A. We leveraged back-translation to expand the training datasets for both monolingual and subtask B. The RoBERTa-base and XLM-R models, enhanced by the MPU framework, showed improved detection of short texts in both monolingual and multilingual settings. Finally, the stacking method allowed us to combine the strengths of multiple models, improving our system's predictive accuracy. Future efforts will consider incorporating advancements in cross-lingual pre-trained models in our subsequent work to further enhance the model's understanding of texts across diverse languages and cultural backgrounds.

\section*{Acknowledgements}
This work is funded by the Natural Science Foundation of Shandong Province under grant ZR2023QF151.


\bibliography{custom}

\appendix



\end{document}